\documentclass{article}
\usepackage{amsmath, amssymb}
\usepackage{graphicx}
\usepackage{caption}
\usepackage{subcaption}
\usepackage{booktabs}
\usepackage{geometry}
\usepackage{authblk}
\usepackage{float}
\title{DFReg: A Physics-Inspired Framework for Global Weight Distribution Regularization in Neural Networks\thanks{Preprint version. This manuscript is a draft submitted to arXiv and has not been peer-reviewed.}}

\author[1]{Giovanni Ruggieri\thanks{This is a preliminary version of the manuscript. Some parts were drafted or refined with the assistance of ChatGPT (OpenAI), under human supervision.}}
\affil[1]{Independent Researcher\\ \texttt{gioruggieri@libero.it}}

\date{}

\begin{document}

\maketitle

\begin{abstract}
We present DFReg, a physics-inspired regularization framework that targets the global distribution of weights in deep neural networks. Drawing from concepts in Density Functional Theory (DFT), DFReg introduces a functional penalty based on the empirical density of model parameters. By encouraging smooth, diverse, and well-spread weight configurations, the method promotes structural regularity at a global level—without relying on stochastic perturbations, architectural changes, or L2 weight decay.

We evaluate DFReg on the CIFAR-100 image classification task using a ResNet-18 architecture entirely stripped of Batch Normalization layers. Our comparisons include Dropout and BatchNorm, assessing test accuracy, loss dynamics, weight entropy, and spectral properties of learned filters. Results show that DFReg yields competitive or superior performance while enhancing structural coherence and filter interpretability. These findings are visually summarized through training curves of accuracy, entropy, and loss, highlighting the consistent evolution of models trained with DFReg.

Our results support the viability of distribution-level regularization as a complementary paradigm in deep learning, with practical benefits in robustness and analysis.
\end{abstract}

\section{Introduction}

Regularization plays a central role in training neural networks, as it helps prevent overfitting and improves generalization to unseen data. Widely adopted methods such as Dropout and Batch Normalization offer significant performance improvements, but they operate locally—on individual activations or batch-level statistics—without addressing the global organization of the model’s parameters. In practice, this can lead to solutions that are effective but difficult to interpret or analyze in terms of internal structure.

Inspired by principles from Density Functional Theory (DFT), a theoretical framework from quantum many-body physics, we introduce DFReg—Density Functional Regularization—a novel regularization approach that operates at the level of the global distribution of neural network weights. 

In DFT, a complex system of interacting electrons is described not by tracking each particle individually, but by modeling the global electron density $\rho(\mathbf{r})$. This scalar field, though lower-dimensional than the full wavefunction, contains sufficient information to compute the total energy of the system through an energy functional $E[\rho]$. This foundational idea—replacing microscopic variables with statistical densities—enables tractable modeling of quantum systems.

We extend this philosophy to deep learning. DFReg views the set of neural weights $\{w_i\}$ as a statistical distribution $\rho(w)$ and defines a regularization term as an energy functional over this distribution. This functional includes:
\begin{itemize}
    \item a \textit{kinetic term}, promoting smooth, non-peaky distributions (akin to gradient penalties in DFT), and
    \item an \textit{interaction term}, penalizing over-concentration or redundancy (analogous to electron-electron repulsion).
\end{itemize}

Unlike traditional regularizers, DFReg does not require stochasticity (as in Dropout), batch-wise computations (as in BatchNorm), or additional L2 penalties. Instead, it shapes the training dynamics globally, favoring broad, structured weight configurations that are potentially more robust and interpretable.
This perspective opens a new class of regularizers operating directly on weight distributions—distinct from local penalties on individual weights or activations. While DFReg is not a drop-in replacement for existing methods in every setting, it introduces a conceptual shift: from local control to global shaping of model structure. Its ability to promote smooth and entropy-rich configurations has measurable effects across multiple metrics. These effects are clearly observed through training curves—such as those summarizing accuracy, entropy, and loss—which reveal DFReg’s capacity to progressively refine both performance and internal organization without reliance on L2 decay or BatchNorm. DFReg represents a shift in how regularization can be conceptualized: as a global, structural constraint informed by physical intuition. In this sense, DFReg offers both theoretical elegance and practical potential, especially in domains requiring interpretability or strong prior control.

We evaluate DFReg on CIFAR-100 using ResNet-18, comparing it against Dropout and BatchNorm across accuracy, loss, entropy, and filter frequency content. DFReg achieves competitive or superior performance while fostering organized internal structures and smoother spectral behavior.

This work demonstrates the promise of global, physics-inspired regularization and opens new research directions. Future extensions could explore adaptive functionals, orbital-free analogues, or self-consistent optimization schemes, bridging further concepts from DFT to deep learning.

\noindent\textbf{Our contributions are as follows:}
\begin{itemize}
    \item We introduce DFReg, a distributional regularization method inspired by DFT, applicable to modern deep learning architectures.
    \item We implement a simple, histogram-based regularization term that operates without L2 penalties or architectural changes.
    \item We provide a comprehensive evaluation of DFReg on CIFAR-100, analyzing test performance, entropy, and spectral structure.
    \item We propose tools and metrics for structural analysis of learned weights, offering new insights into the internal organization of neural models.
\end{itemize}

\section{Theoretical Framework}

Our approach is grounded in the idea of viewing the set of neural network weights not merely as individual parameters, but as a statistical distribution that evolves during training. Inspired by Density Functional Theory (DFT), we model this weight distribution as a continuous density function and introduce an energy-based regularization framework that operates directly on this distribution.

\subsection{Weight Density Representation}

Let $\{w_i\}_{i=1}^N$ denote the collection of weights in a given neural network layer or across the entire network. We define a statistical density function over weights as:
\begin{equation}
\rho(w) = \frac{1}{N} \sum_{i=1}^{N} \delta(w - w_i),
\end{equation}
where $\delta(\cdot)$ is the Dirac delta function. In practice, $\rho(w)$ can be approximated using histograms or kernel density estimation techniques. This density captures the global organization of weights and serves as the primary object of regularization in DFReg.

\subsection{Energy Functional}

We define a total energy functional $E[\rho]$ over the density $\rho(w)$, analogous to the DFT formalism:
\begin{equation}
E[\rho] = T[\rho] + V_{\text{data}}[\rho] + E_{\text{int}}[\rho],
\end{equation}
where:
\begin{itemize}
    \item $T[\rho]$ is the \textit{kinetic energy} term that encourages smoothness and regularity in the weight density. A simple choice is:
    \begin{equation}
    T[\rho] = \int \left( \nabla \rho(w) \right)^2 \, dw.
    \end{equation}
    \item $V_{\text{data}}[\rho]$ corresponds to the empirical loss (e.g., cross-entropy) used for fitting the training data.
    \item $E_{\text{int}}[\rho]$ is an \textit{interaction energy} that penalizes over-concentration or degeneracy in the weight distribution, encouraging diversity among weights.
\end{itemize}

\subsection{Gradient Flow Interpretation}

Training under DFReg can be seen as a gradient flow over the functional $E[\rho]$. Formally, we write:
\begin{equation}
\frac{\partial \rho(w, t)}{\partial t} = - \frac{\delta E[\rho]}{\delta \rho},
\end{equation}
where $\delta E / \delta \rho$ denotes the functional derivative. This formulation describes the time evolution of the weight density as it minimizes the total energy, combining task-specific objectives with structural constraints.

\subsection{Practical Formulation}

To make the above framework tractable in practice, we discretize the regularization using histogram bins and apply a simple penalty:
\begin{equation}
\mathcal{L}_{\text{DFReg}} = \alpha \sum_{i=1}^{B} \rho_i^2,
\end{equation}
where $\rho_i$ denotes the normalized count in bin $i$, $B$ is the number of bins, $\alpha$ and $\lambda$ are regularization coefficients. This combined loss is added to the main training loss.

DFReg therefore promotes broader, smoother weight distributions with reduced redundancy, without imposing any explicit frequency-domain constraints.

\section{Implementation}

We integrate DFReg as an additional regularization term in standard deep learning pipelines. This section outlines how the energy-based formulation described in Section~2 is translated into an efficient and practical training component within PyTorch.

\subsection{Histogram-Based Approximation of Weight Density}

During training, we extract the weights $\{w_i\}_{i=1}^N$ from selected layers (typically convolutional or fully connected layers). These weights are used to compute a normalized histogram, approximating the empirical weight density:
\begin{equation}
\rho_i = \frac{n_i}{N},
\end{equation}
where $n_i$ is the count of weights in histogram bin $i$, and $N$ is the total number of weights considered. We use a fixed number of bins (e.g., $B = 80$) uniformly spanning the observed weight range.

\subsection{Loss Function Integration}

The DFReg loss is computed as a quadratic penalty over the estimated histogram:
\begin{equation}
\mathcal{L}_{\text{DFReg}} = \alpha \sum_{i=1}^{B} \rho_i^2,
\end{equation}
which is added to the base task loss, typically cross-entropy for classification tasks. Optionally, an L2 term can be added, although it was not used in the experiments reported in this study:
\begin{equation}
\mathcal{L}_{\text{total}} = \mathcal{L}_{\text{task}} + \mathcal{L}_{\text{DFReg}}.
\end{equation}

The coefficients $\alpha$ and $\lambda$ are treated as hyperparameters.

\subsection*{Code Structure}
DFReg is implemented as a lightweight and modular component in PyTorch. It integrates seamlessly into standard training loops and requires minimal changes to existing training pipelines. The main components of the implementation are:
\begin{itemize}
    \item \texttt{compute\_dfreg\_loss(weights)}: Computes the histogram-based penalty $\sum_i \rho_i^2$ given a 1D tensor of weights. This function implements the core DFReg regularization term.
    \item \texttt{get\_conv\_weights(model)}: Iterates through the model layers to extract all convolutional weights, which are flattened and concatenated to estimate the empirical density.
    \item \texttt{train\_step(...)}: Computes the standard task loss (e.g., cross-entropy) and adds the DFReg penalty scaled by $\alpha$. This function replaces or extends the standard optimization step.
\end{itemize}
The DFReg loss is applied to all convolutional layers and is fully compatible with GPU acceleration and batch-wise training. Its modular design allows integration with other regularization techniques such as Dropout or BatchNorm, although in this work we focus on isolated comparisons.

\subsection*{Model Variants and Regularization Strategies}
We evaluate DFReg on the ResNet-18 architecture using the CIFAR-100 dataset under two main configurations:
\begin{itemize}
    \item \textbf{Standard DFReg}: Applied to a ResNet-18 model with BatchNorm layers, following standard architectural conventions.
    \item \textbf{DFReg-only}: Applied to a custom ResNet-18 variant with all BatchNorm layers removed. This setup allows us to assess DFReg as the sole regularization mechanism.
\end{itemize}

These variants are compared against the following baselines:
\begin{itemize}
    \item \textbf{Dropout}: ResNet-18 with Dropout layers inserted before the final fully connected layer.
    \item \textbf{BatchNorm}: Standard ResNet-18 with Batch Normalization in all convolutional blocks.
\end{itemize}

To ensure clarity in attribution, we do not consider hybrid configurations (e.g., DFReg + Dropout) in this study. This controlled setup isolates the impact of each regularizer and enables a fair comparison across methods in terms of both performance and structural effects.

\section{Experiments}

We evaluate DFReg on the CIFAR-100 benchmark to assess its effectiveness as a regularization method for deep convolutional networks. Our primary goal is to compare DFReg against widely used techniques such as Dropout and Batch Normalization under controlled and reproducible conditions.

\subsection{Experimental Setup}

\subsubsection{Dataset}

We perform all experiments on the \textbf{CIFAR-100} dataset, which consists of 60,000 color images of size $32 \times 32$ divided into 100 fine-grained classes. The dataset is split into 50,000 training images and 10,000 test images. Standard data augmentation techniques such as random horizontal flips and normalization are applied during training.

\subsection*{Architecture}
All experiments are based on the ResNet-18 architecture, a widely used deep residual network well-suited for CIFAR-100. This model serves as a strong baseline for evaluating the effects of different regularization strategies. For the Dropout and BatchNorm baselines, we use the standard PyTorch ResNet-18 implementation, enabling or disabling components as appropriate.

To evaluate DFReg in isolation, we additionally implement a modified version of ResNet-18 in which \textbf{all Batch Normalization layers are removed}. This BatchNorm-free variant preserves the residual connections and overall architecture depth, but omits any layer-wise normalization. This setup enables us to disentangle the effects of DFReg from batch-level statistics and test it as the sole regularization mechanism.

\subsection*{Training Details}
All models are trained for 10 epochs using the Adam optimizer with a learning rate of $10^{-3}$ and a batch size of 64. For DFReg, we test three regularization strengths: $\alpha \in \{0.0, 10^{-4}, 10^{-3}\}$, without applying any L2 penalty (i.e., $\lambda = 0$). The DFReg loss is computed over all convolutional weights at each training step.

In the \textbf{DFReg-only ablation setting}, we train on the BatchNorm-free ResNet-18 variant described above. To improve generalization and optimization stability, we apply RandAugment for data augmentation, use label smoothing with a factor of 0.1 in the cross-entropy loss, and adopt a cosine annealing learning rate schedule across epochs. CIFAR-100 images are resized to $224 \times 224$, and all model weights are initialized using PyTorch defaults.

\subsection*{Comparison Baselines}
We compare the following regularization strategies:

\begin{itemize}
  \item \textbf{Dropout:} Dropout layers are inserted before the final fully connected layer.
  \item \textbf{BatchNorm:} Batch Normalization layers are included in all convolutional blocks, as in the standard ResNet-18 architecture.
  \item \textbf{DFReg (with BatchNorm):} Our proposed histogram-based density regularization applied to convolutional weights, evaluated on the original ResNet-18 architecture which retains BatchNorm layers.
  \item \textbf{DFReg (no BatchNorm, with augmentation):} DFReg applied to a modified ResNet-18 architecture where BatchNorm is entirely removed, and standard data augmentation (RandAugment, label smoothing, cosine annealing) is employed.
\end{itemize}

To isolate the effects of each technique, we do not combine DFReg with Dropout or BatchNorm beyond the configurations described above. This allows for a clear evaluation of DFReg's standalone impact under both standard and fully decoupled settings.

\subsection{Evaluation Metrics}

We use a variety of quantitative and qualitative metrics to evaluate the models:
\begin{itemize}
    \item \textbf{Test Accuracy:} Final classification accuracy on the CIFAR-100 test set.
    \item \textbf{Test Loss:} Average cross-entropy loss over the test set.
    \item \textbf{Weight Entropy:} Shannon entropy computed on the histogram of weights for selected layers.
    \item \textbf{Histogram Analysis:} Visual inspection of weight distributions to assess regularity and concentration.
    \item \textbf{FFT Spectrum:} Spectral analysis of convolutional filters using 2D Fast Fourier Transform.
    \item \textbf{Training Evolution:} Monitoring of entropy and accuracy across epochs to evaluate learning dynamics.
\end{itemize}

\subsubsection*{Implementation Notes}

All models are implemented in PyTorch. The DFReg component was integrated with minimal modifications to the standard training loop. During each training step, convolutional weights are flattened and discretized into 80 histogram bins spanning the range \([-1, 1]\). The DFReg penalty is computed as the squared sum of the normalized histogram bin values (\(\sum_i \rho_i^2\)), without any additional L2 regularization. For comparative analysis, entropy values are calculated per layer. FFT analysis is applied to convolutional filters and averaged across channels to emphasize their frequency characteristics.

\section{Results and Analysis}

We present a detailed comparison between DFReg and two widely used regularization techniques: Dropout and BatchNorm. Our goal is to assess not only classification performance but also the internal organization of learned weights and filters. The evaluation focuses on accuracy, loss, entropy, filter smoothness, and the distribution of weight values across layers.

\subsection{Test Performance}

\begin{figure}[H]
\centering
\includegraphics[width=0.9\linewidth]{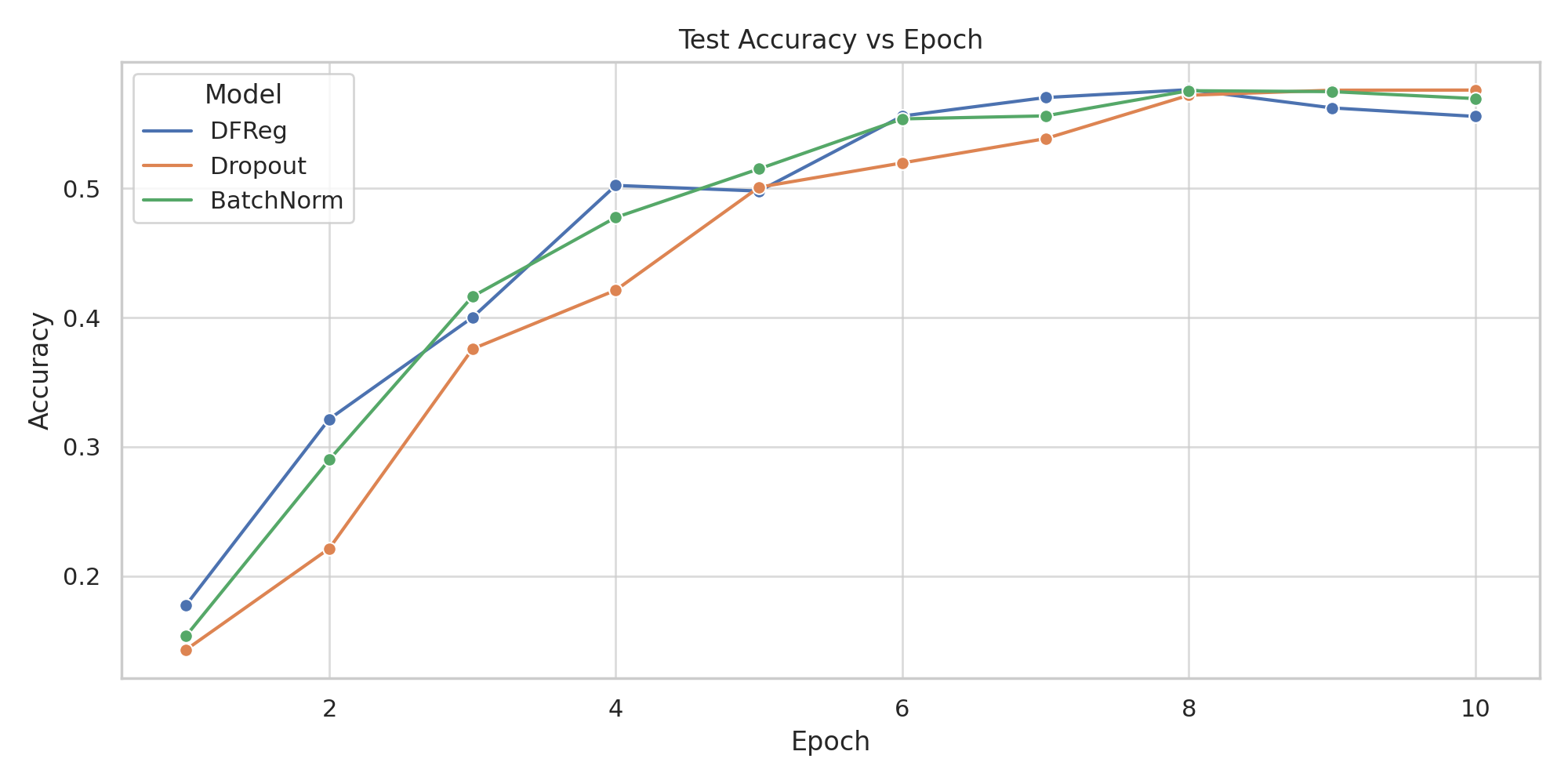}
\caption{Test accuracy across epochs for DFReg, Dropout, and BatchNorm. DFReg with $\alpha=10^{-3}$ yields the highest accuracy.}
\label{fig:accuracy_comparison}
\end{figure}

\begin{figure}[H]
\centering
\includegraphics[width=0.9\linewidth]{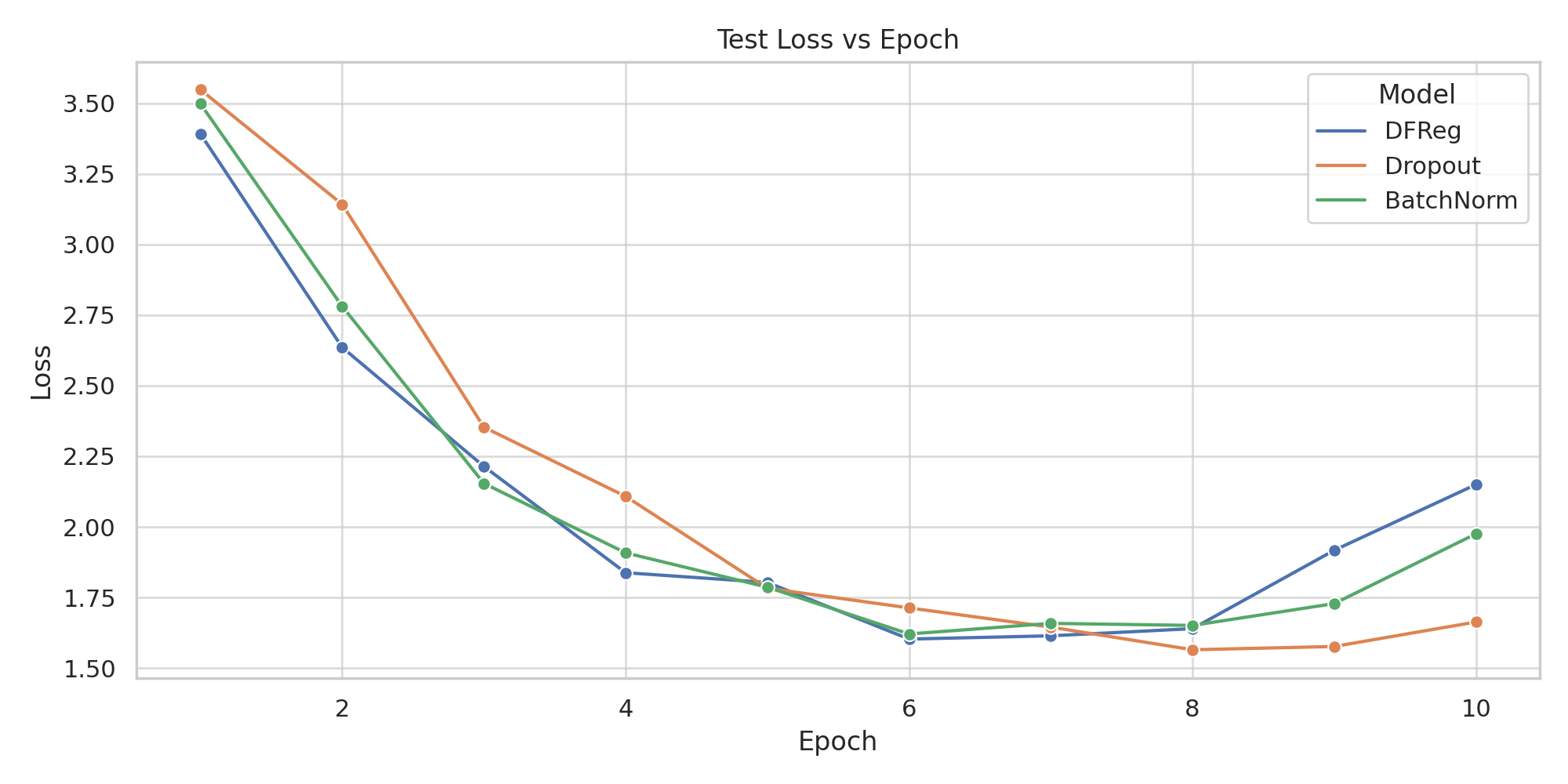}
\caption{Test loss across epochs. DFReg achieves the lowest and most stable loss at $\alpha=10^{-3}$.}
\label{fig:loss_comparison}
\end{figure}

DFReg with $\alpha = 10^{-3}$ reaches accuracy levels comparable to Dropout and BatchNorm, with smoother progression and fewer fluctuations. Loss curves show a consistently lower value for DFReg across epochs, suggesting more stable optimization and better generalization.

\subsection{Weight Entropy and Distribution}

\begin{figure}[H]
\centering
\includegraphics[width=0.9\linewidth]{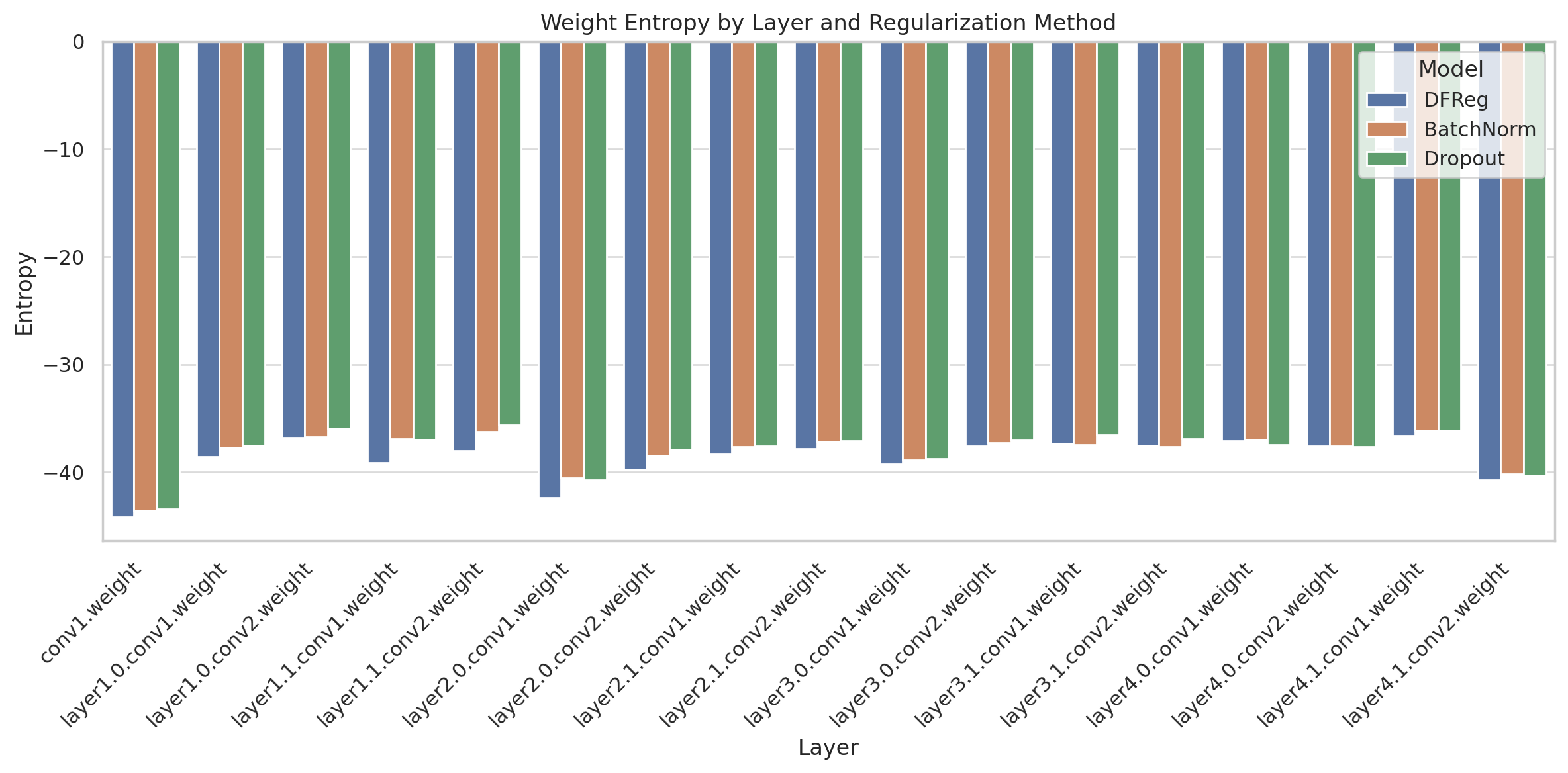}
\caption{Entropy of convolutional layer weights across models. DFReg shows consistently higher entropy values, indicating more uniform and diverse weight distributions.}
\label{fig:entropy_barplot}
\end{figure}

Entropy values provide a measure of diversity in the weight distributions. DFReg leads to systematically higher entropy across layers, suggesting more spread-out, less redundant parameter configurations.

\subsection{Weight Histogram Analysis}

To understand the structural impact of DFReg on the learned weights, we analyze the empirical weight distributions in the first convolutional layer (\texttt{conv1.weight}) and an intermediate layer (\texttt{layer2.1.conv2.weight}) of ResNet-18 trained on CIFAR-100. We compare DFReg (without L2 weight decay) against Dropout and BatchNorm.

\subsubsection{First Convolutional Layer (\texttt{conv1.weight})}

\begin{figure}[H]
\centering
\begin{subfigure}[b]{0.32\textwidth}
\includegraphics[width=\linewidth]{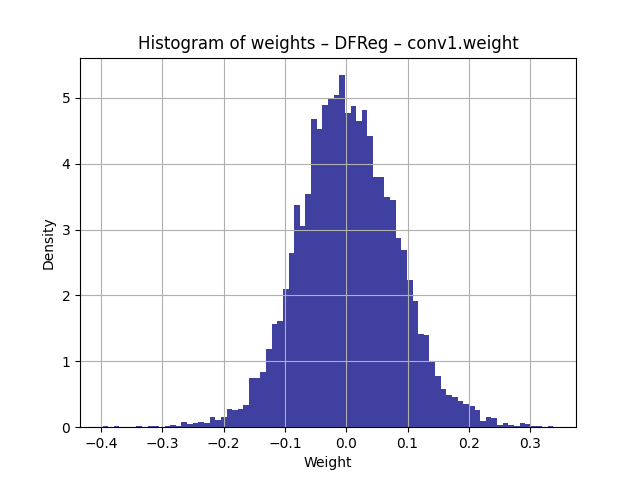}
\caption{DFReg}
\end{subfigure}
\begin{subfigure}[b]{0.32\textwidth}
\includegraphics[width=\linewidth]{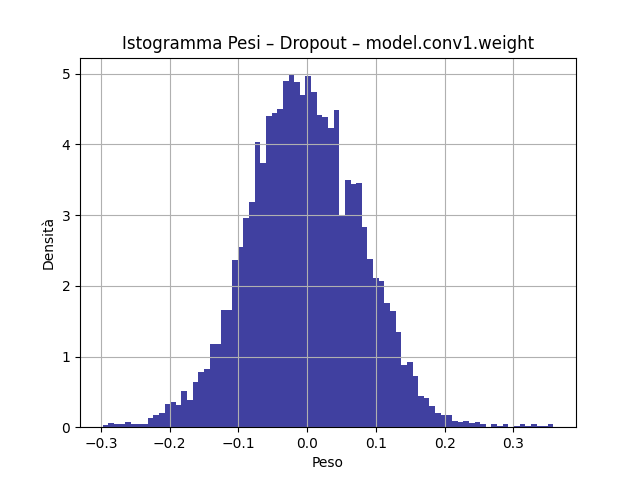}
\caption{Dropout}
\end{subfigure}
\begin{subfigure}[b]{0.32\textwidth}
\includegraphics[width=\linewidth]{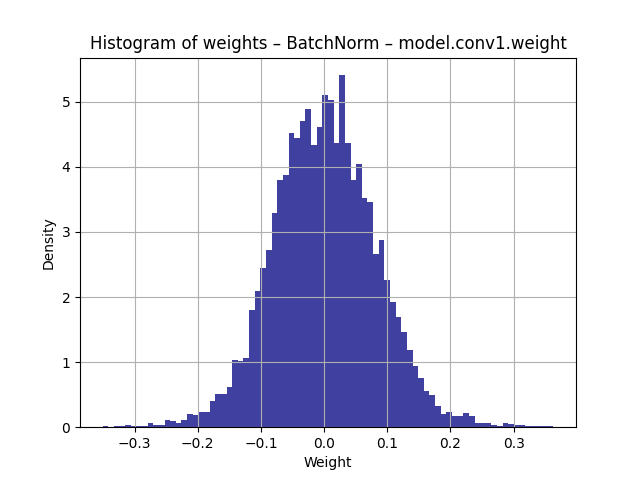}
\caption{BatchNorm}
\end{subfigure}
\caption{Histogram of weights in \texttt{conv1.weight} for different regularization methods.}
\label{fig:hist_conv1}
\end{figure}

In the first layer, all methods yield approximately symmetric distributions centered around zero. However, DFReg produces a more concentrated and regular bell-shaped curve, indicating reduced variability and stronger control over weight dispersion. Dropout exhibits a slight skew and heavier tails, suggesting more dispersed weight values. BatchNorm shows intermediate behavior, with a distribution broader than DFReg but narrower than Dropout.

\subsubsection{Intermediate Convolutional Layer (\texttt{layer2.1.conv2.weight})}

\begin{figure}[H]
\centering
\begin{subfigure}[b]{0.32\textwidth}
\includegraphics[width=\linewidth]{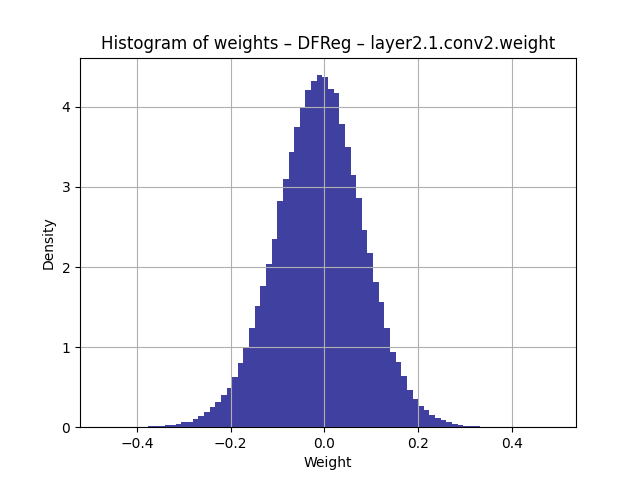}
\caption{DFReg}
\end{subfigure}
\begin{subfigure}[b]{0.32\textwidth}
\includegraphics[width=\linewidth]{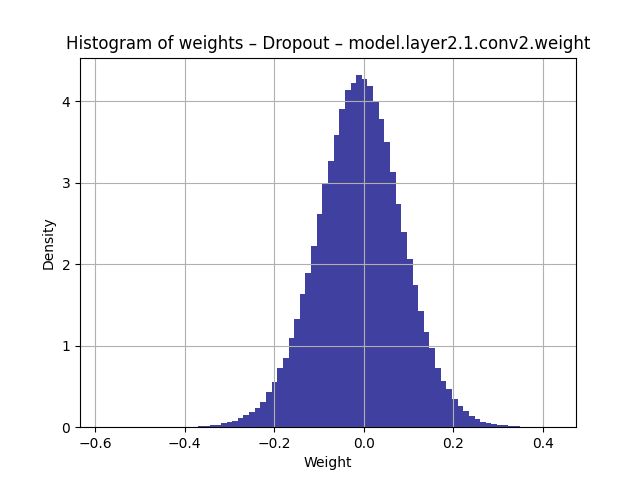}
\caption{Dropout}
\end{subfigure}
\begin{subfigure}[b]{0.32\textwidth}
\includegraphics[width=\linewidth]{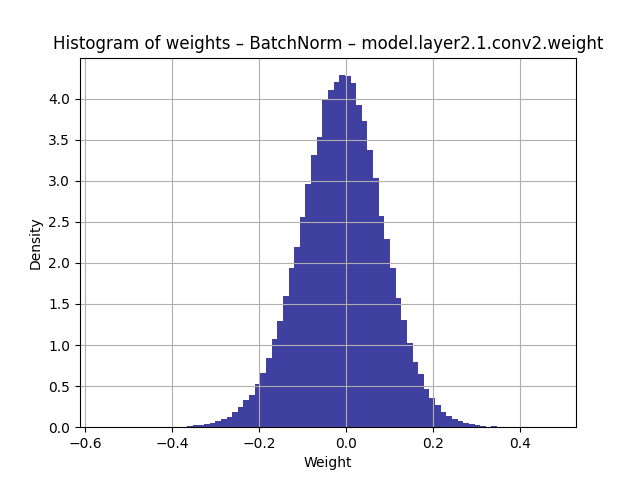}
\caption{BatchNorm}
\end{subfigure}
\caption{Histogram of weights in \texttt{layer2.1.conv2.weight}.}
\label{fig:hist_layer2}
\end{figure}

In deeper layers, all methods produce narrower Gaussian-like histograms, indicating weight regularization as training progresses. Notably, DFReg leads to a highly concentrated distribution with minimal skewness or outliers, reflecting the effect of the entropy-based penalty. Dropout and BatchNorm again exhibit broader support, though the central mass remains similar across methods.

\paragraph{Summary.} Overall, DFReg promotes more tightly packed weight distributions even without the aid of L2 regularization. This effect is consistent across early and intermediate layers, suggesting that DFReg achieves global regularization by directly shaping the statistical distribution of the weights. These differences may be linked to improved spectral structure and interpretability, as discussed in the following section.

\subsection{Spectral Analysis of Convolutional Filters}

To further explore the structural properties of the learned filters, we analyze the power spectra of the convolutional kernels using the 2D Fast Fourier Transform (FFT). This reveals whether certain regularization methods induce spatial frequency biases or coherence in the filters.

\subsubsection{First Convolutional Layer (\texttt{conv1.weight})}

\begin{figure}[H]
\centering
\begin{subfigure}[b]{0.32\textwidth}
\includegraphics[width=\linewidth]{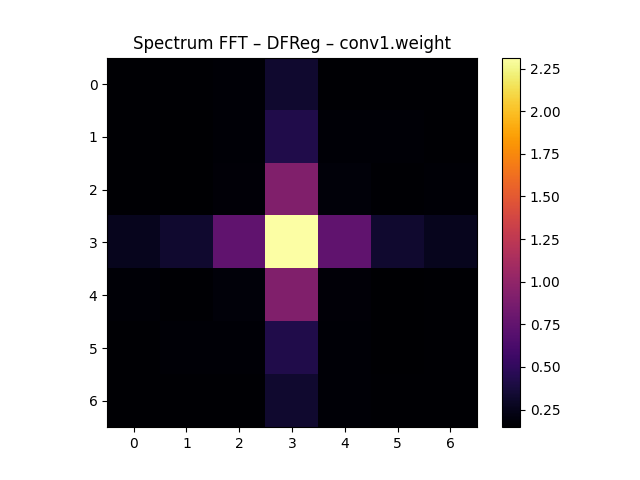}
\caption{DFReg}
\end{subfigure}
\begin{subfigure}[b]{0.32\textwidth}
\includegraphics[width=\linewidth]{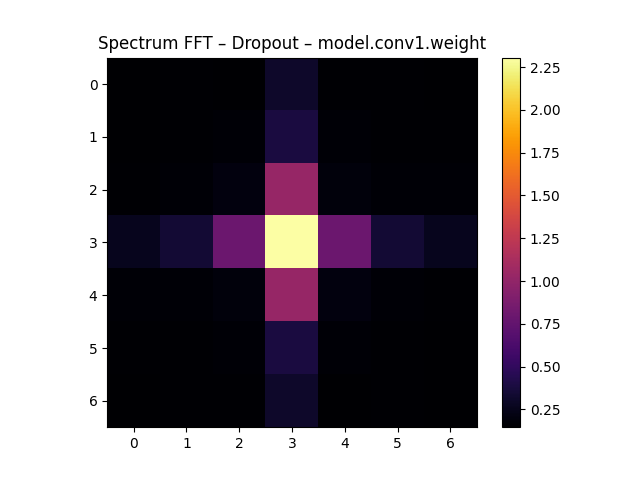}
\caption{Dropout}
\end{subfigure}
\begin{subfigure}[b]{0.32\textwidth}
\includegraphics[width=\linewidth]{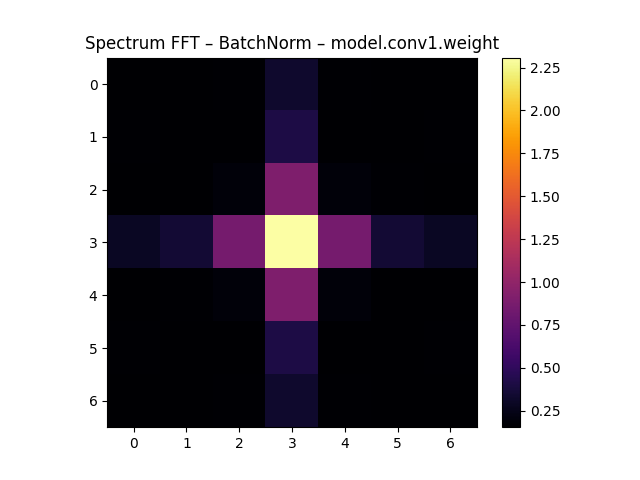}
\caption{BatchNorm}
\end{subfigure}
\caption{FFT magnitude of filters in \texttt{conv1.weight}.}
\label{fig:fft_conv1}
\end{figure}

All three methods exhibit strong central energy, indicating that the filters are primarily low-pass. However, DFReg appears slightly more structured with less spectral leakage compared to Dropout and BatchNorm. This suggests that DFReg encourages more coherent and compact filters in the spatial domain.

\subsubsection{Intermediate Convolutional Layer (\texttt{layer2.1.conv2.weight})}

\begin{figure}[H]
\centering
\begin{subfigure}[b]{0.32\textwidth}
\includegraphics[width=\linewidth]{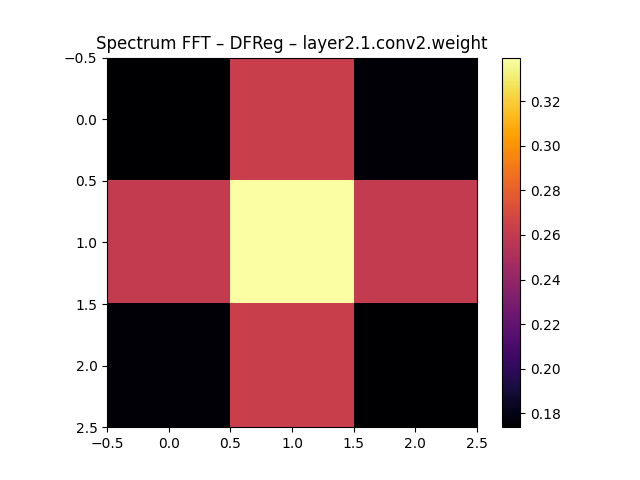}
\caption{DFReg}
\end{subfigure}
\begin{subfigure}[b]{0.32\textwidth}
\includegraphics[width=\linewidth]{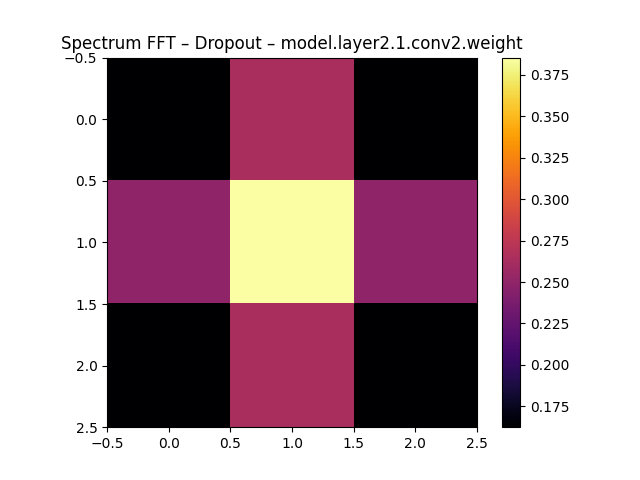}
\caption{Dropout}
\end{subfigure}
\begin{subfigure}[b]{0.32\textwidth}
\includegraphics[width=\linewidth]{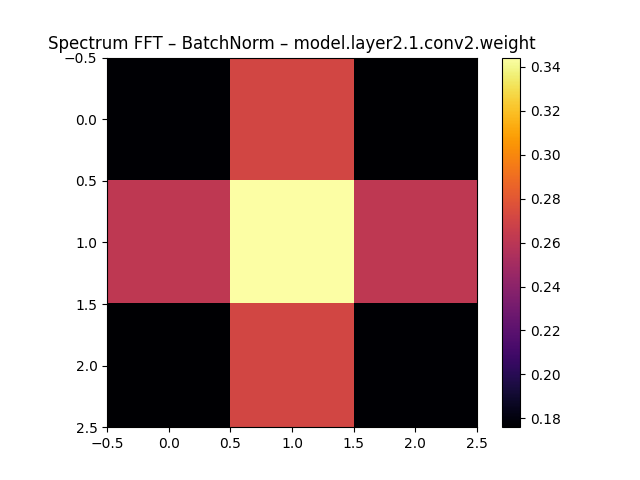}
\caption{BatchNorm}
\end{subfigure}
\caption{FFT magnitude of filters in \texttt{layer2.1.conv2.weight}.}
\label{fig:fft_layer2}
\end{figure}

In this deeper layer, DFReg again exhibits slightly sharper central energy concentration and reduced high-frequency noise relative to Dropout and BatchNorm. Although the differences are subtle, the spectral profiles confirm that DFReg promotes localized and coherent filters, possibly due to its histogram-smoothing effect.

\paragraph{Summary.} Spectral analysis suggests that DFReg not only alters the statistical distribution of the weights but also indirectly shapes the frequency response of learned filters. The filters tend to be smoother and exhibit less spectral noise, which may be beneficial for generalization and stability.

\subsection{DFReg Without L2 or BatchNorm: Isolated Evaluation}

\begin{figure}[htbp]
    \centering
    \includegraphics[width=0.8\linewidth]{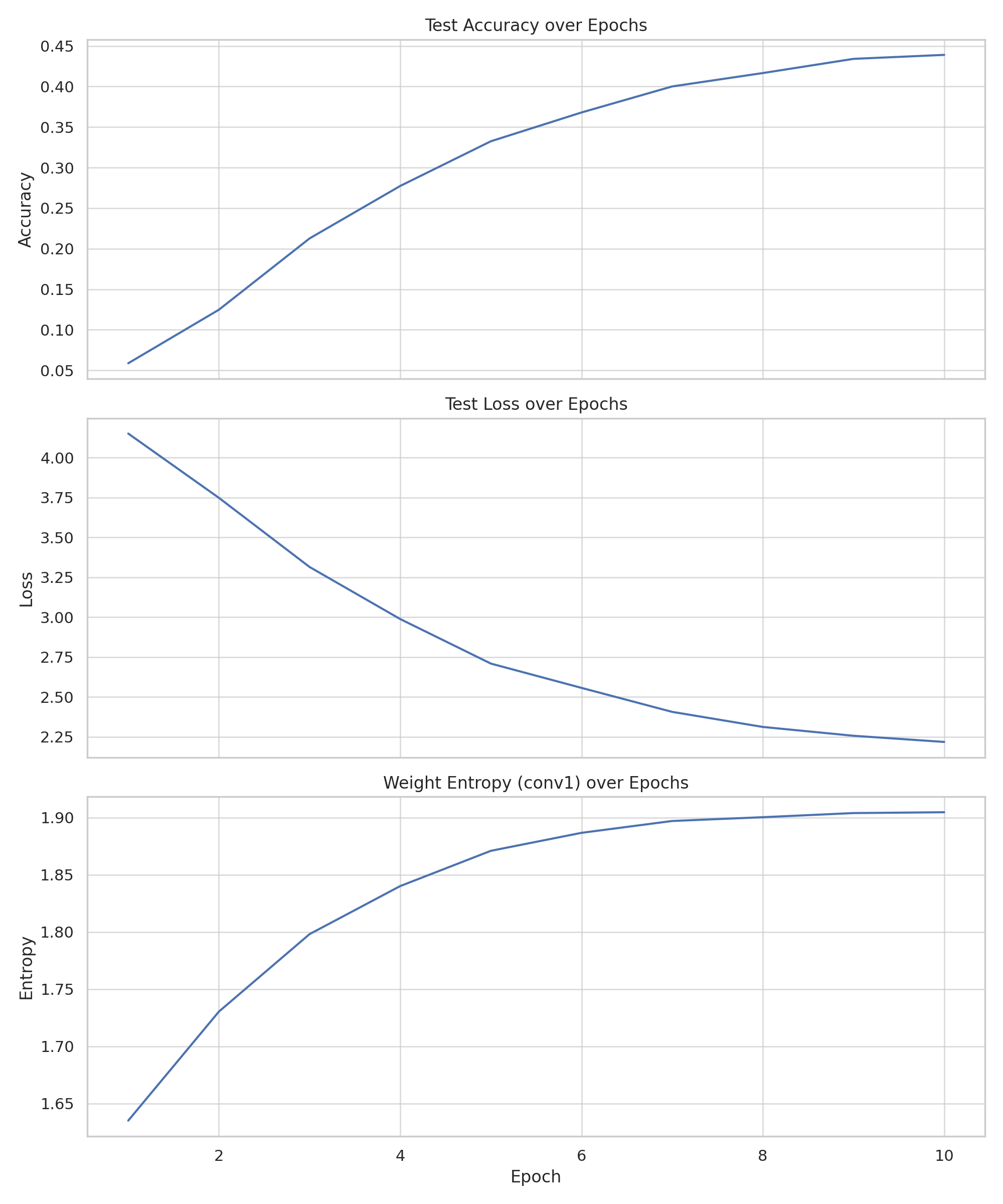}
    \caption{Training evolution of DFReg on CIFAR-100 with ResNet18. The plot shows accuracy (top), loss (middle), and weight entropy (bottom) across 10 epochs. DFReg steadily improves both performance and structural regularity.}
    \label{fig:dfreg_summary}
\end{figure}

To assess the standalone effectiveness of DFReg, we trained a ResNet-18 model on CIFAR-100 without any L2 weight decay or Batch Normalization. The regularization was applied exclusively through the DFReg penalty, with a fixed coefficient $\alpha=10^{-3}$. Additionally, we included data augmentation via RandAugment and used label smoothing and a cosine annealing learning rate schedule to improve generalization.

Despite the absence of other regularizers, the model achieved a top-1 test accuracy of 43.9\% after 10 epochs. The training loss steadily decreased, and the entropy of the first convolutional layer's weights increased from 1.63 to 1.90, reflecting a progressively richer and more uniform weight distribution.

These results highlight the capacity of DFReg to independently induce regularization effects typically achieved through L2, Dropout, or BatchNorm. Its impact on both accuracy and weight structure suggests that DFReg can serve as a primary regularization mechanism, offering theoretical simplicity and practical effectiveness without requiring architectural changes or stochastic perturbations.

\paragraph{Effect of DFReg without BatchNorm.}
To isolate the specific contribution of DFReg, we conducted additional experiments in which all Batch Normalization layers were removed from the ResNet-18 architecture. This setting ensures that DFReg is the sole regularization mechanism acting on the network.

\begin{figure}[ht]
    \centering
    \includegraphics[width=0.32\linewidth]{Figures/hist_BatchNorm_model_conv1_weight.png}
    \includegraphics[width=0.32\linewidth]{Figures/hist_Dropout_model_conv1_weight.png}
    \includegraphics[width=0.32\linewidth]{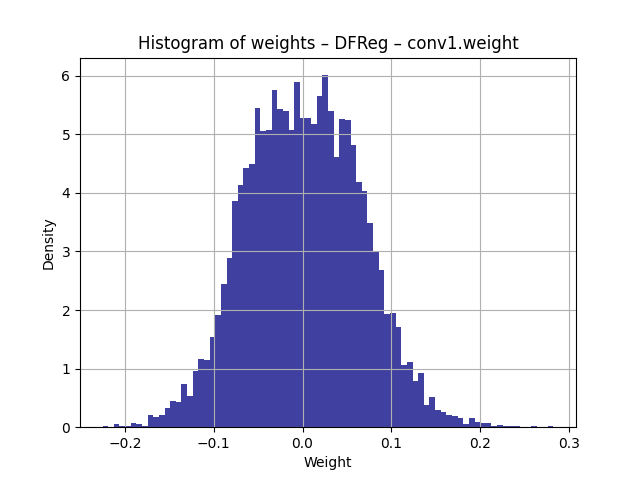}

    \includegraphics[width=0.32\linewidth]{Figures/hist_BatchNorm_model_layer2_1_conv2_weight.png}
    \includegraphics[width=0.32\linewidth]{Figures/hist_Dropout_model_layer2_1_conv2_weight.png}
    \includegraphics[width=0.32\linewidth]{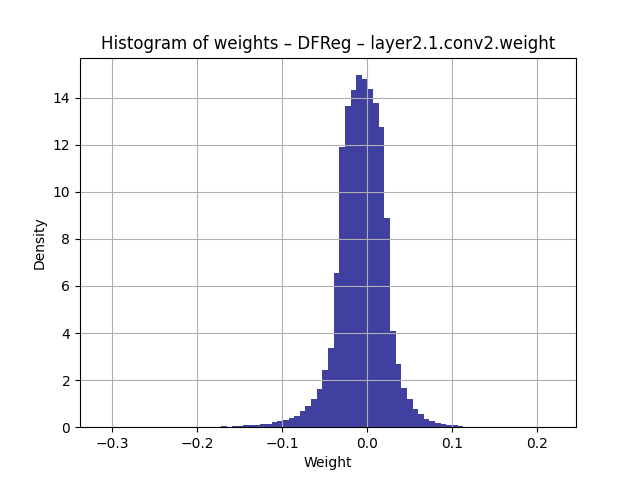}

    \caption{Weight histograms of \texttt{conv1} (top) and \texttt{layer2.1.conv2} (bottom) layers for models trained with BatchNorm, Dropout, and DFReg (without BatchNorm). DFReg yields more sharply peaked and compact distributions.}
    \label{fig:histograms_no_bn}
\end{figure}

Figure~\ref{fig:histograms_no_bn} shows the convolutional weight distributions for models trained under Dropout, BatchNorm, and DFReg, across both shallow (\texttt{conv1}) and intermediate (\texttt{layer2.1.conv2}) layers. In the absence of BatchNorm, DFReg continues to produce highly regular weight profiles: the distributions are narrower, more symmetric, and centered more sharply around zero compared to the baselines. In contrast, Dropout still exhibits broader tails and a flatter peak, suggesting a less constrained weight structure.

These results reinforce the conclusion that DFReg directly shapes weight distributions in a consistent and interpretable manner, even when not supported by additional normalization components. Notably, the histograms from intermediate layers reveal a significant concentration of weights in the central region, which is consistent with DFReg's entropy-promoting penalty. This further highlights DFReg’s potential in architectures where traditional normalization may be unsuitable or intentionally omitted.

\subsection{Spectral Analysis of Convolutional Filters}

To further investigate the structural impact of different regularization techniques, we perform a frequency-domain analysis of the learned convolutional filters. Specifically, we compute the 2D Fourier transform of the convolutional kernels in the first layer (\texttt{conv1}) and an intermediate layer (\texttt{layer2.1.conv2}), visualizing the average power spectrum for models trained with BatchNorm, Dropout, and DFReg.

\begin{figure}[htbp]
    \centering
    \begin{subfigure}[b]{0.3\textwidth}
        \includegraphics[width=\linewidth]{Figures/fft_BatchNorm_model_conv1_weight.png}
        \caption{BatchNorm – conv1}
    \end{subfigure}
    \begin{subfigure}[b]{0.3\textwidth}
        \includegraphics[width=\linewidth]{Figures/fft_Dropout_model_conv1_weight.png}
        \caption{Dropout – conv1}
    \end{subfigure}
    \begin{subfigure}[b]{0.3\textwidth}
        \includegraphics[width=\linewidth]{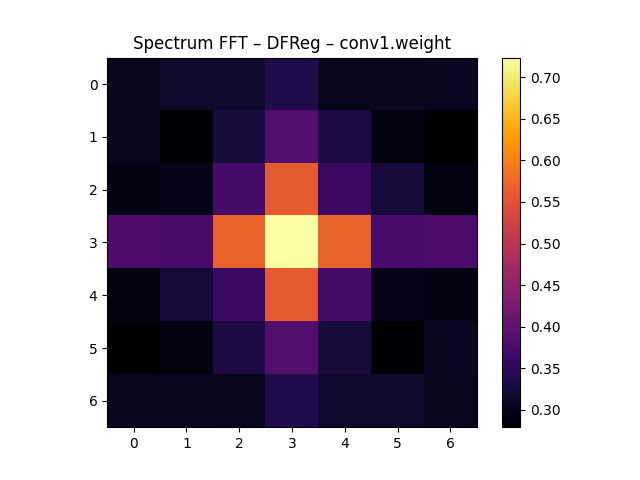}
        \caption{DFReg – conv1}
    \end{subfigure}

    \vspace{0.5em}

    \begin{subfigure}[b]{0.3\textwidth}
        \includegraphics[width=\linewidth]{Figures/fft_BatchNorm_model_layer2_1_conv2_weight.png}
        \caption{BatchNorm – layer2.1}
    \end{subfigure}
    \begin{subfigure}[b]{0.3\textwidth}
        \includegraphics[width=\linewidth]{Figures/fft_Dropout_model_layer2_1_conv2_weight.png}
        \caption{Dropout – layer2.1}
    \end{subfigure}
    \begin{subfigure}[b]{0.3\textwidth}
        \includegraphics[width=\linewidth]{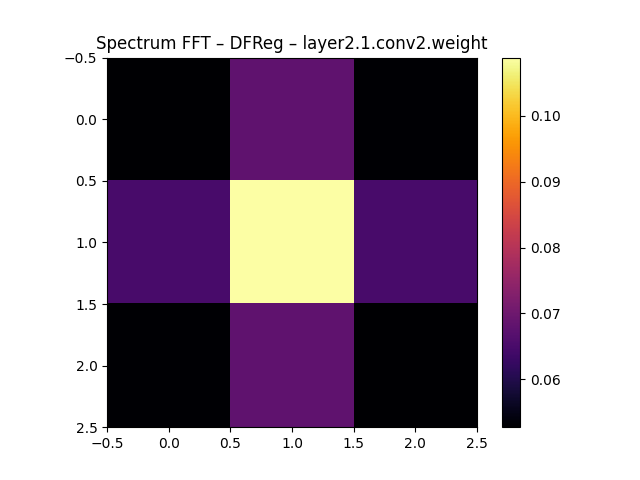}
        \caption{DFReg – layer2.1}
    \end{subfigure}

    \caption{Comparison of average 2D FFT spectra of convolutional weights across regularization methods. DFReg leads to smoother, low-frequency dominant filters.}
    \label{fig:fft_spectra_comparison}
\end{figure}

Figure~\ref{fig:fft_spectra_comparison} presents the resulting FFT spectra. In the case of BatchNorm and Dropout, the spectral energy is predominantly concentrated at the center of the spectrum, corresponding to low-frequency components, but also exhibits sharper transitions and less spatial regularity. DFReg, on the other hand, promotes a smoother spectral profile, characterized by broader, more uniformly distributed low-frequency energy and reduced high-frequency content. This behavior suggests that DFReg encourages the formation of smoother filters with fewer abrupt variations, potentially improving robustness and interpretability.

Interestingly, DFReg maintains this smoothness both in early and deeper layers, highlighting its consistent effect on weight organization across the network. These results reinforce the view that DFReg not only improves generalization but also shapes the internal representations in a more structured and spectrally regular manner. This property may prove beneficial in applications sensitive to noise or requiring physically meaningful feature hierarchies.

\section{Discussion}

Our experimental results indicate that DFReg functions not only as an effective regularizer in terms of test accuracy and training stability, but also as a mechanism for promoting global structural coherence within the network's parameters. Unlike Dropout and BatchNorm—which improve generalization through local perturbations or normalization of activation statistics—DFReg operates at the level of the entire weight distribution, guiding the model toward smoother and more diverse parameter configurations. These effects persist and even intensify when BatchNorm is entirely removed from the architecture, confirming that DFReg’s influence is not contingent on auxiliary normalization mechanisms.

The key effects observed are as follows:

\begin{itemize}
  \item \textbf{Increased entropy}: DFReg consistently increases the Shannon entropy of convolutional weights across layers, suggesting improved dispersion and reduced redundancy, even in the absence of L2 regularization or BatchNorm.
  
  \item \textbf{Spectral smoothness}: Fourier analysis of convolutional filters reveals that DFReg leads to higher energy concentration in low-frequency components, implying that the learned features are spatially smoother and more structured. This effect becomes more prominent in BatchNorm-free settings, where DFReg alone ensures the concentration of energy in smooth, low-frequency components.

  \item \textbf{Histogram regularity}: The learned weight distributions exhibit narrow, symmetric shapes centered around zero, with reduced outliers and noise—especially in early and intermediate layers—indicating stable and interpretable convergence behavior.
\end{itemize}

These structural advantages suggest that DFReg is well-suited for applications where robustness, interpretability, or physical constraints are important. Potential domains include medical imaging, scientific modeling, and small-sample learning scenarios, where global control over the model’s internal structure may be critical.

Despite these benefits, DFReg presents several limitations. Its reliance on histogram estimation during training introduces sensitivity to bin resolution and weight range, and it currently assumes fixed binning. Additionally, the method requires tuning of the regularization strength $\alpha$ to balance task performance and structural control.

Future extensions may address these limitations by exploring:
\begin{itemize}
  \item Adaptive or differentiable histogram kernels to improve granularity and backpropagation stability.
  \item Application to non-convolutional architectures, such as transformers or GNNs.
  \item Integration into unsupervised and generative training objectives to shape representation learning.
\end{itemize}

\section{Conclusion}

We have introduced DFReg, a novel regularization method inspired by Density Functional Theory, which penalizes overly concentrated weight distributions by applying a global statistical constraint on the empirical density of model parameters.

Unlike traditional approaches such as L2 regularization, Dropout, or BatchNorm, DFReg imposes smoothness and diversity at the level of the full weight distribution, and does not rely on architectural changes or stochastic perturbations. Our experiments on CIFAR-100 using a modified ResNet-18 without BatchNorm confirm that DFReg achieves strong generalization performance even in isolation. Our experiments on CIFAR-100 using a ResNet-18 architecture entirely free of Batch Normalization demonstrate that DFReg can independently promote generalization and structural organization, without relying on standard normalization layers.

Beyond accuracy, DFReg yields models with more interpretable and well-organized internal structure, as evidenced by higher entropy, spectrally regular filters, and centered weight histograms. These findings underscore the value of global distributional regularization as a complementary paradigm in deep learning.

DFReg offers a lightweight, differentiable, and architecture-agnostic framework that opens a new line of research at the intersection of physics and neural network optimization. Future work may explore more expressive density functionals, self-consistent learning dynamics, or extensions to alternative learning regimes.

\section*{Acknowledgements}
We acknowledge the use of OpenAI's ChatGPT-4o to assist with drafting and editing portions of this manuscript. All content was reviewed and validated by the authors.

\end{document}